\documentclass{article}
\usepackage{spconf_ieee,amsmath,graphicx}
\usepackage{xcolor}
\usepackage{amsfonts}
\usepackage{amssymb,amsmath,bm}
\usepackage{multirow}
\usepackage{booktabs}
\usepackage{multicol}
\usepackage{array}
\usepackage{enumitem}
\usepackage{multirow, makecell}
\usepackage[mathscr]{euscript}
\usepackage{float}
\usepackage{hyperref}
\usepackage{tablefootnote}
\usepackage{graphics}
\usepackage{cite}

\newcommand{\myparagraph}[1]{\vspace{.4em} \noindent \textbf{#1}\ }

\ninept
\title{Maestro-U: Leveraging joint speech-text representation learning for zero supervised speech ASR}
\name{
Zhehuai~Chen,~Ankur~Bapna,~Andrew~Rosenberg,~Yu~Zhang, \\
Bhuvana~Ramabhadran,~Pedro~Moreno,~Nanxin~Chen \thanks{Thanks to Gary Wang, Jesse Emond, Charles Yoon, Zhong Meng and Kevin Hu  for many discussions and infratructure related assistance.
}\thanks{Copyright 2023 IEEE. Published in the 2022 IEEE Spoken Language Technology Workshop (SLT) (SLT 2022), scheduled for 19-22 January 2023 in Doha, Qatar. Personal use of this material is permitted. However, permission to reprint/republish this material for advertising or promotional purposes or for creating new collective works for resale or redistribution to servers or lists, or to reuse any copyrighted component of this work in other works, must be obtained from the IEEE. Contact: Manager, Copyrights and Permissions / IEEE Service Center / 445 Hoes Lane / P.O. Box 1331 / Piscataway, NJ 08855-1331, USA. Telephone: + Intl. 908-562-3966.}}
\address{Google, Inc.}

\begin{document}
%\ninept
%
\maketitle
\begin{abstract}

Training state-of-the-art Automated Speech Recognition (ASR) models typically requires a substantial amount of transcribed speech.  In this work, we demonstrate that a modality-matched joint speech and text model introduced in~\cite{zhehuai2021} can be leveraged to train a massively multilingual ASR model without any supervised (manually transcribed) speech for some languages. %In most zero resource conditions, lack of transcribed speech also implies lack of lexicons.
This paper explores the use of jointly learnt speech and text representations in a massively multilingual, zero supervised speech, real-world setting to expand the set of languages covered by ASR with only unlabeled speech and text in the target languages. Using the FLEURS dataset, we define the task to cover $102$ languages, where transcribed speech is available in $52$ of these languages and can be used to improve end-to-end ASR quality on the remaining $50$. First,  we  show that by combining speech representations with byte-level text representations and use of language embeddings,  we  can  dramatically  reduce  %the  supervised speech  requirements for developing  ASR systems in new languages. On the FLEURS dataset, this approach is able to reduce 
the Character Error Rate (CER) on languages  with  no  supervised speech from  64.8\% to 30.8\%, a relative reduction of 53\%.  Second, using a subset of South Asian languages we show that Maestro-U can promote knowledge transfer from languages with supervised speech even when there is limited to no graphemic overlap. Overall, Maestro-U closes the  gap  to  oracle  performance  by  68.5\%  relative and reduces the CER of 19 languages below 15\%.  %with the target languages, reducing the average CER of the target languages from 100.0 to 18.3 .  We believe this is the first demonstration that  competitive  ASR  performance  can  be  achieved  for  an unseen language using no language resources other than text and untranscribed speech.

\end{abstract}
\begin{keywords}
Speech-text Representation learning, Zero Resource, Massively Multilingual zero-supervised-speech  ASR 
\end{keywords}
\section{Introduction}
\label{sec:intro}

\begin{figure*}[t]
  \centering
\includegraphics[width=0.7\linewidth]{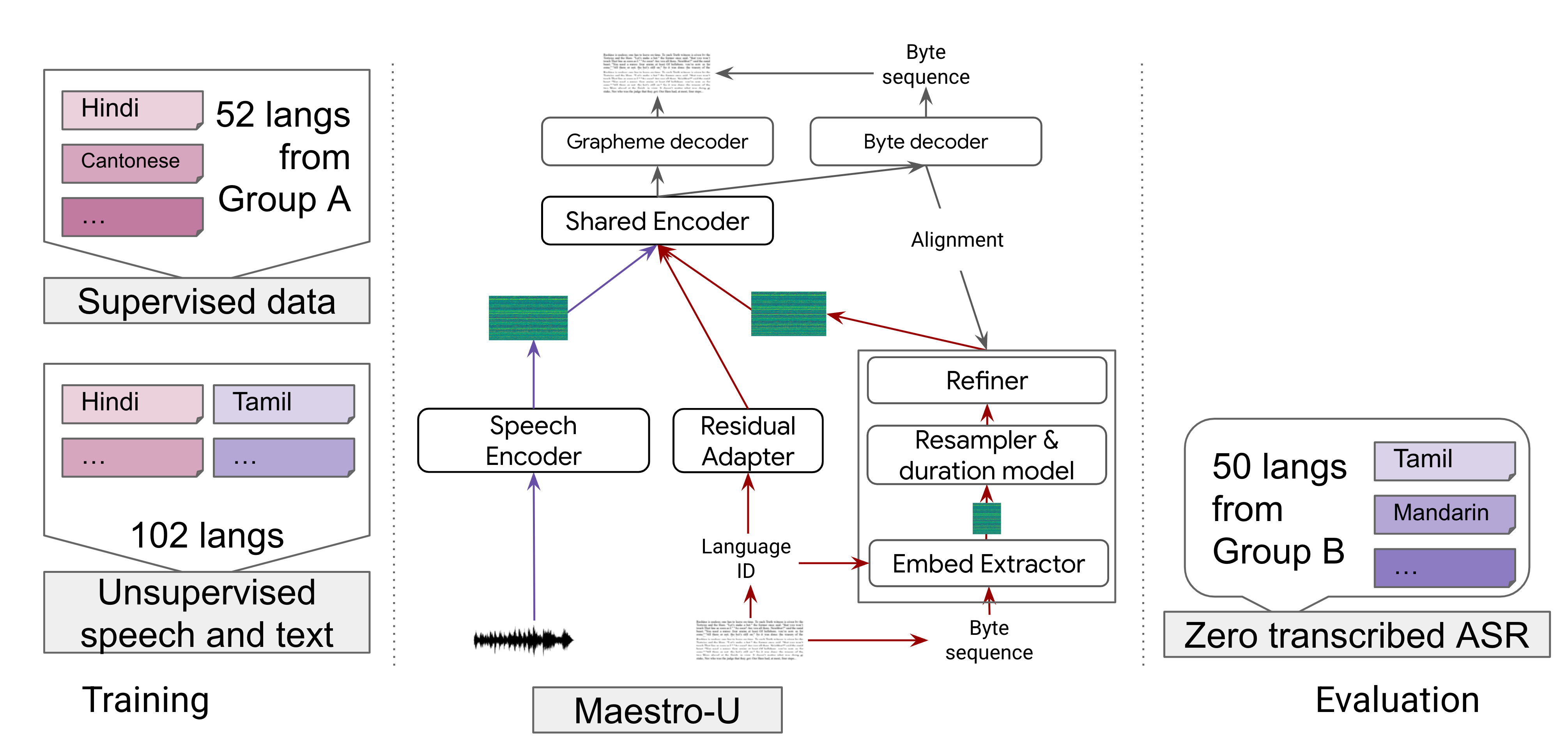}
  %https://docs.google.com/presentation/d/1tMTIPJCW7tnExVOY0T7Cbvee-Wp587f8GFRyuQcY600/edit#slide=id.gf492a22dda_0_3
    \caption{Maestro-U Training for Zero Supervised Speech ASR. 
    %%added for camera-ready - begin
    The zero supervised speech ASR  is defined in Section~\ref{sec:benchmark}. Speech, shared and text encoders are described in Section~\ref{sec:maestro}. The use of bytes, language ID and residual adapter  are described in Section~\ref{sec:mlang-design} and \ref{sec:unseen}. 
    %%added for camera-ready - end
    }
    \label{fig:framework}
\end{figure*}

\copyrightnotice{978-1-6654-7189-3/22/\$31.00~\copyright2023 IEEE}

The last few years have seen the emergence of two major directions of research towards improving low resource ASR quality. The first direction uses  multilingual models to leverage the large amounts of supervised (manually transcribed) speech available for high resource languages to improve quality on low resource languages~\cite{heigold2013multilingual,cho2018multilingual,pratap2020mls,li2022massively}. The second direction utilizes self-supervised pre-training on large amounts of unlabeled speech~\cite{oord2018representation,chung2018speech2vec,schneider2019wav2vec,baevski2020wav2vec}, unlabeled text~\cite{bahdanau2016end,kannan2018analysis} or both~\cite{bapna2021slam,bapna2022mslam,chen2022maestro} to complement the relatively small amounts of transcribed data available for these languages. An extreme example of the low-resource setting is learning ASR without the availability of any (in-language) transcribed resources (\emph{zero-supervised-speech } ASR). In this work, we explore the possibility of using jointly learnt speech and text representations~\cite{bapna2022mslam,chen2022maestro} to expand ASR to languages lacking transcribed speech resources.

%{Any other zero-resource / unsupervised paper we want to mention here. Maybe the Zero-Speech challenges? 2019 focus on TTS without text from Speech while 2017 is more related. Also wav2vec-U? There is also 2021 which was to learn LMs from acoustcis only}
The zero-supervised-speech  setting has previously been explored in several works~\cite{liu2018completely,yeh2018unsupervised,chen2019completely,baevski2021unsupervised}. However, most prior research on unsupervised ASR either learns models for phoneme recognition (implicitly assuming a model for phoneme to grapheme conversion), or assumes the availability of grapheme to phoneme (G2P) models for text augmentation.  The construction of a G2P model requires at least as much expert human knowledge and effort as speech transcription. As such they are unavailable for many of the worlds' languages. In many zero resources settings, lack of a lexicon can double the unit error rate~\cite{liu2022towards}.
In the ZeroSpeech2021~\cite{nguyen2020zero} challenge, researchers explored the ability of models to learn language models with raw speech and no textual resources. These models were evaluated on their ability to learn phonetics, lexicon, syntax and semantic structures in the language.

In this work, we define a practical setting in line with real world constraints, assuming the availability of unlabeled speech and  text (graphemes) in all $102$ languages under consideration, and the availability of supervised speech in $52$ of these languages. Given these resources, we attempt to improve end-to-end ASR quality on the remaining $50$ zero-supervised-speech  languages. We establish that a joint speech-text representation learning model, Maestro~\cite{chen2022maestro} fails to perform well on this zero supervised speech task, reaching an average Character Error Rate (CER) of $54.2\%$ averaged over $50$ languages. %(Section \ref{sec:exp-ablation}). 
% forward reference
To improve the joint speech and text representation learning for this setting we propose the following:
%\begin{itemize}
%    \item Language embeddings and adapters~\cite{kannan2019large} in pre-training (Section \ref{sec:mlang-design})
%   \item Replacing graphemes with Byte-level text representations (Section \ref{sec:unseen}) and,
%   \item  Tapering of  self-supervised losses
%\end{itemize}
%resulting in a final zero-supervised-speech  average CER of $29.6\%$. 

%The main contributions of this paper are:
\begin{itemize}[leftmargin=1em]
\item Building on the FLEURS benchmark~\cite{conneau2022fleurs}, we define a massively multilingual zero-supervised-speech  ASR task motivated by real-world constraints, with the goal of expanding the set of languages covered by ASR models.  
\item We propose several improvements to the Maestro described in~\cite{zhehuai2021}, namely, the use of  language embeddings and adapters to learn better mappings (Section \ref{sec:mlang-design}) across speech and text in languages sharing writing systems; and use of byte level text representations to enable better transfer to script-unique zero-supervised-speech  languages (Section \ref{sec:unseen}). 
%These improvements also help on the low resource fine-tuning setting.
\item We analyze and compare the role of different text injection strategies, including using phonemized text and byte-level text representations %text and transliterated text 
to understand the role of shared vocabularies in zero-supervised-speech  ASR (Section~\ref{sec:unseen})
\item We conduct ablations of components used in representation learning to understand the role of our proposed techniques and those proposed in\cite{chen2022maestro}, including the importance of the learnt duration model and consistency losses.
\end{itemize}
The proposed work in this paper results in a final zero supervised speech  average CER of $30.8\%$, a relative reduction of 43\% relative over Maestro ~\cite{zhehuai2021}. %53\%. %43 comes from 54.2-30.8
To the best of our knowledge, we believe this is the first demonstration that  competitive  ASR  performance  can  be  achieved  for  an unseen language using no language resources other than unspoken text and untranscribed speech.

\section{FLEURS Zero Supervised Speech ASR}
\label{sec:benchmark}

We define our massively multilingual zero supervised speech ASR task building on the FLEURS benchmark~\cite{conneau2022fleurs}. FLEURS is a publically available, multi-way parallel dataset of just $10$ hours of read speech in each of the $102$ languages spanning $7$ geo-groups, which can be used as a benchmark task for ASR. Of the $102$ languages present in the FLEURS benchmark, we choose $52$ to serve as our supervised languages (\emph{Group A}) while the remaining $50$ will be utilized in a zero-supervised-speech  setting (\emph{Group B}). In order to understand zero supervised speech  performance across all geo-groups, we balance the number of languages in Groups A and B from each geo-group as shown in Table~\ref{tab:fleurs_langs}. %. The list of languages from the Fleurs benchmark in Group A and Group B are given in Table~\ref{tab:fleurs_langs}.

\begin{table*}[ht]
    %\centering
    \caption{Languages in Group A and Group B from the Fleurs benchmark.}
    \label{tab:fleurs_langs}
    
    \begin{tabular}{ll}
    \toprule
    \textit{Group  A}\\ 
    \midrule
     \textbf{Central Asian, Middle-East and North Africa (CMN)}: Azerbaijani (az), Kazakh (kk) \\
    Kyrgyz (ky), Mongolian (mn), Pashto (ps), Persian (fa) \\
         \textbf{CJK}: Cantonese (yue), Korean (ko) \\
         \textbf{Eastern European (EE)}: Belarusian (be), Estonian (et), Georgian (ka), Latvian (lv), Macedonian (mk), Polish (pl), Slovak (sk), Serbian (sr) \\
         \textbf{South Asia (SA)}: Bengali (bn), Hindi (hi), Malayalam (ml), Nepali (ne), Punjabi (pa), Urdu (ur), Telugu (te) \\
         \textbf{South-east Asia (SEA)}: Cebuano (ceb), Indonesian (id), Khmer (km), Malay (ms), Thai (th), Vietnamese (vi) \\
         \textbf{Sub-Saharan Africa (SSA)}:  Amharic (am), Hausa (ha), Kamba (kam), Lingala (ln), Northern-Sotho (nso), Oromo (om), Somali (so) \\
         Swahili (sw),  Wolof (wo), Yoruba (yo) \\
         \textbf{Western European (WE)}: American English (en), Bosnian (bs), Croatian (hr), Finnish (fi), French (fr), German (de), Greek (el), Irish (ga) \\ 
         Italian (it), Kabuverdianu (kea), Maltese (mt), Occitan (oc), Welsh (cy)\\
        \midrule
     \textit{Group B} \\ 
     \midrule
     \textbf{CMN}: Arabic (ar), Hebrew (he), Sorani-Kurdish (ckb), Tajik (tg), Turkish (tr), Uzbek (uz) \\
    \textbf{CJK}: Japanese (ja), Mandarin (cmn) \\
    \textbf{EE}: Armenian (hy), Bulgarian (bg), Czech (cs), Lithuanian (lt), Romanian (ro), Russian (ru), Slovenian (sl), Ukrainian (uk) \\
    \textbf{SA}: Assamese (as), Gujarati (gu), Kannada (kn), Marathi (mr), Oriya (or), Sindhi (sd), Tamil (ta) \\
    \textbf{SEA}: Burmese (my), Filipino (fil), Javanese (jv), Lao (lo), Maori (mi) \\
    \textbf{SSA}: Afrikaans(af), Fula (ff), Ganda (lg), Igbo (ig), Luo (luo), Nyanja (ny), Shona (sn), Umbundu (umb), Xhosa (xh), Zulu (zu) \\
    \textbf{WE}: Asturian (ast), Catalan (ca), Danish (da), Dutch (nl), Galician (gl), Hungarian (hu), Icelandic (is), Latin American Spanish (es) \\
    Luxembourgish (lb), Norwegian (nb), Portuguese (pt), Swedish (sv) \\
    \bottomrule
    \end{tabular}
    \vspace{-3mm}
\end{table*}

In addition to FLEURS, following ~\cite{bapna2022mslam,chen2022maestro}, we also include supervised speech and unlabeled speech from the MLS~\cite{pratap2020mls}, VoxPopuli~\cite{wang2021voxpopuli}, CommonVoice~\cite{ardila2019common} and Babel~\cite{gales2014speech} datasets when available. % and appropriate.% and paired speech corresponding to Group A languages from MLS, VoxPopuli and Babel. 
While mC4~\cite{xue2020mt5} is a good text resource for injection, it contains noisy data that can hurt ASR quality~\cite{kreutzer2022quality}. Therefore, we cleaned this text further using the language-id and wordlist-based approaches described in ~\cite{caswell2020language}.
%For unpaired text, we utilize the mC4 text dataset for our initial experiments. However, as also highlighted in ~\cite{kreutzer2022quality}, we find mC4 to contain noisy text that hurts quality for several zero supervised speech  languages. For our clean text dataset we utilize a proprietary web-mined dataset cleaned using LangID and wordlist-based approaches, following~\cite{caswell2020language}, spanning all FLEURS languages except ast, kea and umb.
%We also study the inclusion of the transcription-only data from FLEURS Group B  to see the effect of {\em in-domain text} data.

To understand  graphemic overlap between the supervised languages $\mathrm{L}^{(A)}$ and zero supervised speech languages $\mathrm{L}^{(B)}$ and its effect on the
%knowledge transfer and
ASR performance, we define the unseen grapheme ratio $\gamma(l)$ of language $l$
in group B 
w.r.t.   group A languages in Equation~\ref{eqn:ugr},
\begin{eqnarray}
\label{eqn:ugr}
&\gamma(l) =1-\frac{ \Big|\cup_{k=1}^n \big(V(l) \cap V(\mathrm{L}^{(A)}_k)\big)\Big|}{|V(l)|},\ l\in \mathrm{L}^{(B)} %\\
%&\gamma(\mathrm{L}^{(B)}) = \sum_{k=1}^{n}{\gamma(l_k)}/n,\ l_k\in \mathrm{L}^{(B)}
\end{eqnarray}
where $V(l)$ denotes the grapheme vocabulary of the language $l$, which can be obtained from any text resource for $l$. In this work, we obtain  $V(l)$ from the FLEURS release in~\cite{conneau2022fleurs}.

\section{Proposed Method: Maestro-U}
\label{sec:maestro-u}
%Maestro review and describe how to use maestro in zero-rsc
%\bhuv this section needs a rewrite.
In this work, we pursue the idea of expanding an ASR model to new languages while requiring zero supervised speech, using only text and untranscribed speech.  
This is done by text injection with the previously proposed Maestro~\cite{chen2022maestro} with a series of innovations to handle unseen scripts and promote multilingual knowledge transfer. Figure~\ref{fig:framework} summarizes the Maestro-U training process.

\subsection{Text injection using Maestro}
\label{sec:maestro}
%{Motivate our solution}  % copy them as below
Zero- or few-shot approaches require training models that can map one sequence to another implicitly. This has been achieved for several text style transfer and MT tasks via training cross-lingual models with GANs \cite{lee2019cross}, self-supervised pre-training and for ASR for mapping audio to phonemes with GANs \cite{baevski2021unsupervised}. Recent work on speech-text pre-training, like mSLAM \cite{bapna2022mslam} and Maestro~\cite{chen2022maestro}, have demonstrated that it is possible to learn shared representations of speech and text in the same model. 
%\andrew{not sure this next sentence is necessary but leaving for now}
%In this work, we start with investigating Maestro to learn zero supervised resource ASR. 

%review Maestro

Maestro was proposed in~\cite{chen2022maestro} to address the speech-text representation learning problem by first aligning text to speech using an RNN-T decoder and then training a {\it Text Encoder}. The resultant text encoder can be used to map unspoken  text to this aligned shared space and learn from it.
When learning from untranscribed speech data, we use contrastive loss on the speech encoder outputs and a masked language model (MLM) loss on the shared encoder output 
similar to W2v-BERT~\cite{chung2021w2v}.
When learning from paired speech and text, the text encoder uses this RNN-T model to generate alignments between the text targets and the speech encoder output. The {\it Resampler} and {\it Refiner} layers replicate the initially learned text embeddings to match the duration of the speech embedding using this alignment information and a Mean-Squared Error (MSE) training objective is used to enforce consistency between the resultant speech and text representations. 

When learning from unspoken text, speech-text alignment information is unavailable. Therefore, Maestro uses  durations predicted from a duration prediction model in a fashion similar to speech synthesis~\cite{elias2021parallel}. This model is trained %on any available paired data 
to predict the duration of each token. The predicted duration on unspoken text is subsequently used to upsample the learned text embeddings to match the speech frame-rate.
%The resultant  speech-aligned and resampled text embeddings can be learnt by enforcing RNN-T loss with the unpaired text as the target
RNN-T loss is applied over the resultant upsampled text embeddings with masking in frequency and time domain similar to SpecAugment~\cite{park2019specaugment}. This allows for the use of the same RNN-T objective on both speech embeddings or text embeddings.

%text representation to match a W2v-BERT speech representation. 
%Maestro learns unified representations through sequence alignment, duration prediction and  matching embeddings in the learned space through an aligned masked-language model loss.

%how to apply
When applied in the zero supervised speech ASR, the text duration model  used to upsample the text representation to the shared speech-text space  is trained on only those languages where transcribed speech is available. This duration model is used to upsample the text representation of every language whether or not paired data is available. 
This approach allows us to use this upsampled text representation to train an ASR model purely on unspoken text from an unseen language.

%Maestro improvements for 
\subsection{Promote multilingual knowledge transfer}
\label{sec:mlang-design}
%Lang id and adapters
We found that Maestro works reasonably well for several languages (Section \ref{sec:exp-overall}), but sometimes recognizes audio from zero resource languages with graphemes from supervised languages. This phenomenon is common in many multilingual ASR systems~\cite{datta2020language}. 
% since it has not been trained to map audio to text in zero-supervised-speech languages. 
To encourage the model to generate text in the correct language, we modify Maestro to utilize language-id signals and language-specific parameters in several ways.

Given that new languages and their scripts are learned only through text data, we inject a learned language embedding in the input of the RNN-T decoder to bias the output script. % of the model similar to~\cite{cite a MT paper?}.  
Additionally, language information can be missing when using a common text representation across languages (e.g. phonemes) (Section~\ref{sec:unseen}). Thus, we add the language information back to the text encoder with a similar learned language embedding. 
The shared encoder of the Maestro model is the backbone to learn joint speech-text representations.
To introduce language dependent parameters there, we 
 use language id conditioned residual adapter layers~\cite{kannan2019large} to condition the shared encoder.  Residual adapters are small feed forward networks typically two layers with a bottleneck dimension (though other structures are possible) whose inputs are the inputs to a conformer block and whose outputs are added to the output of the same block. Language-id conditioned residual adapters are applied to all conformer blocks in the shared encoder.  %We find using adapter layers to substantially improve the multilingual performance when training without any paired data.

% Moreover, the language condition also helps the Maestro model associate the unpaired speech representations and unpaired text representations.

%Loss scaling of text and w2b
The Maestro model in~\cite{chen2022maestro} relies on the supervised fine-tuning after the Maestro training process to get the best performance. In  zero supervised speech ASR,  there is no supervised data in the target language for fine-tuning. We will show in Section~\ref{sec:exp-overall} that supervised fine-tuning on  Group A languages can hurt the zero-transcribed performance on Group B. 
%\{Additional improvements from using clean and in-domain data and loss tapering.}
%text loss; self supervised loss
In this zero supervised speech setting, we empirically found two additional techniques to improve performance of languages in Group B: i) upscaling unspoken text loss weight to 12.0; ii) loss tapering of the W2v-BERT loss in the last 50k steps of the training. These two modifications focus the training on text sources. 
% We believe it stems from the fact that the only way to learn to produce textual form of the target languages in this setting is through the text injection.
An ablation study describes the value of these modifications in Section~\ref{sec:exp-ablation}.

%Define zero-rsc setting ; fleurs
%How it is more realistic for lang expansion?
% bhuv this section needs a rewrite clearly mentioning the methods particularly the projection of utf-8 down to a 256 dim embedding
%\andrew{This section needs to be restructured.  Start with the graphemic representation and its limitatinos, then the byte representation. the phoneme and translit are both high resource approaches that may be availabel for some languages but not all.}

\begin{table*}[tb]
\caption{\label{tab:exp-all} {Overall performance of Maestro-U: Models trained in the {\it supervised} setting include the available supervised training data (10 hours per language), while models trained in the {\it zero supervised training data} setting do not include any supervised data from the 50 languages in group \textbf{B}. } }
\centering
\begin{tabular}{l|l|l|l|l|l|l|l}
\toprule
method                                       & Paired  & Text & Speech & Finetune& \multicolumn{3}{c}{CER}               \\
 \midrule
\multicolumn{5}{l|}{{\em Supervised setting on languages in Group A and B}} & \textbf{A+B}& A &B \\
 \midrule
 %rerun eval https://paste.googleplex.com/5728802664611840?raw
 %corresponding WER eval result (still pretty off) https://paste.googleplex.com/5846916949606400?raw
 %FtSpeechOnly600MBaseFleursb4.d-00023400 14.5
 %Finetune600MBaseTemp1Rnnt.h-00022000 12.9844
 W2v-BERT                                     & A+B           & N    & Y  &Y    & \textbf{12.3}   & 11.8     & 12.8 \\
%W2v-BERT                                     & A+B           & N    & Y  &Y    & 13.0   & 11.38      & 14.6 \\
% didn't finish. remove https://paste.googleplex.com/5893765479792640
%Standard Maestro                              & A+B           & Y    & Y &Y     & TODO                     & TODO        & TODO                     \\
%FtFleursMaestroFleursAdpBe.h 
Maestro-U (proposed)                             & A+B           & Y    & Y    &Y  & \textbf{8.7} & 7.8    & 9.7  \\
%Proposed Maestro                             & A+B           & Y    & Y    &Y  & {8.9}  & 7.6      & 10.4 \\
 \midrule
\multicolumn{5}{l|}{{\em Zero supervised resource setting on languages in Group B}} &A+B& A &\textbf{B}      \\
 \midrule
%nosp2.FtSpeechOnly600MBaseFleursGroupA.y-00024000 0.661872 50 0.413383 102 0.174452 52
W2v-BERT                             & A           & N    & Y  &Y    & 41.3   & 17.4     & \textbf{66.2}  \\
%W2v-BERT                             & A           & N    & Y  &Y    & 40.4  & 16.8      & 65.0 \\
 %nosp2.NoTextMlsBabelVoxpMaestroFleursA.i-00077000 0.647636 50 0.465006 102 0.2894 52
Joint W2v-BERT  & A             & N    & Y    &N  & 46.5                   & 29.0      & \textbf{64.8}  \\
%Joint W2v-BERT  & A             & N    & Y    &N  & 45.5                   & 27.6      & {64.1} \\
% the result is too bad. removed. more info: https://docs.google.com/document/d/1hBoJOCID7MnkFBZTcDjcFFDW7tFbfGv4HbSx3wxMHQM/edit?resourcekey=0--1BWR-VwN6wxk044F4ISCA#heading=h.bnzejqmtdad0
%\ \ \ \ + LM fusion                            & -             &    - &- &       - & TODO                     & TODO        & TODO                     \\
%nosp2.MC4MlsBabelVoxpMaestroFleursA.i-00037000 0.542154 50 0.395733 102 0.254944 52 11
Standard Maestro                              & A           & Y    &Y& N      & 39.6                  & 25.5      & \textbf{54.2}                   \\
%Standard Maestro                              & A           & Y    &Y& N      & 37.7                     & 21.3        & 54.8                     \\
%M4MlsBabelVoxpMaestroFleursA8LidACleanTbAdapterBe.k M4MlsBabelVoxpMaestroFleursAaLidACleanTbAdapterBe.k
%nosp2.M4MlsBabelVoxpMaestroFleursAaLidACleanTbAdapterBe.k-00107000 0.30812 50 0.211365 102 0.117812 52 13
Maestro-U (proposed)                             & A             & Y    &Y& N      & {21.1}  & 11.8     & \textbf{30.8}  \\
%Proposed Maestro                             & A             & Y    &Y& N      & {20.5} & 11.8      & {29.6} \\
 \bottomrule
\end{tabular}
\end{table*}

\subsection{Handling unseen writing systems (scripts)}
\label{sec:unseen}

%describe the problem
Our preliminary text injection experiment shows that performance of this approach substantially suffers when the target Group B language (being trained without any transcribed speech) has a textual form (script) (for example Brahimic), that has no overlap with any  language in Group A, where transcribed speech is available.  
When the text encoder, trained on Group A, has never observed a script, the reliability of the alignments and thus the shared representation predictions suffer. 
The central problem we need to solve here is how to share information across scripts.  We do this by converting input graphemes (text) into a common representation that is shared across all languages.

%solution:
%Design 1: text encoder input
%Promote knowledge sharing between languages: byte, phone, etc. 
%Expensive: G2P
%Moderate: transliteration
%Cheap: byte (deterministic clustering writing space) 

We explore different text representations for the input to the Maestro text encoder to enable more overlap across languages. 
Notably, while deriving these varieties of intermediate representations, the output targets (a.k.a. RNN-T decoder output) of the Maestro model is always the same native script grapheme units.

%\andrew{first}
Assuming no extra human knowledge, we look at whether the model can learn  script clustering implicitly.
Since we assume that the text is machine readable, there is necessarily a digital representation of the text. 
%We use the UTF-8 byte coding of the input text to establish a shared 256-dim text representation for the text encoder.
We use the UTF-8 byte coding of the input text to establish a shared text representation of 256 encoding vocabulary size  for the text encoder.
%\andrew{second}

Another approach is to use a pronunciation model that can map input text to phoneme (IPA, X-SAMPA) sequences, thereby modeling phonetic representations in the text encoder.  These phonetic representations describe speech sounds in a language agnostic inventory. In this way, the duration and text representation prediction can be shared and implicitly clustered across different language scripts. However, high quality pronunciation models are expensive, and not readily available for all of the world's languages, which limits the utility of this method.

A third approach to this problem might be to transliterate the text into a different script \cite{emond2018transliteration}. However, it is not clear that transliteration is a simpler problem to solve than pronunciation modeling.  Transliteration models  as they can be as error prone as pronunciation models.  While we consider transliteration a reasonable approach to address the problem of unseen scripts, we leave this direction to future work.

% Given our target is to expand language supports to low-resource languages, we investigate alternative methods requiring less human knowledge than phoneme annotation. 
% Transliteration models~\cite{emond2018transliteration} are somewhat less resource intensive than pronunciation models, but still require expert knowledge of the target language. As an example,  we can use latin-script transliteration of the Cyrillic text as the text encoder input but still predict the Cyrillic as the model output. The other disadvantage is that human transliteration annotation is usually non-deterministic, which introduces extra complexity in the modeling.
%
% This improves performance for languages where graphemes are shared with existing supervised languages; but for new languages with a large fraction of unseen graphemes the model has never seen a mapping from acoustics - graphemes. To resolve this issue, we experiment with sharing the `text' vocabulary across languages - 2 approaches: (i) lightweight requiring low linguistic resources - byte level vocabs, (ii) requiring additional linguistic resources - using g2p models for phoneme level text injection. We find that phoneme level injection performs the best, but using a simple byte level vocab is not far behind.

% emphasize how we use byte, which is similar to phoneme in prev maestro
We include these intermediate representations in a manner similar to \cite{chen2022maestro}. We include  two RNN-T decoders to model bytes and graphemes respectively. The byte RNN-T is then used to obtain the alignment for the text encoder when learning with bytes as  input. 
With this design, the unspoken text is first converted to byte encodings and then extracted and upsampled by the text encoder. The text representations are then fed to the shared encoder. Both byte and grapheme decoders are used to enforce unspoken text learning through RNN-T  losses. The final ASR hypothesis is produced by the target grapheme RNN-T decoder.
%Finally, the grapheme RNN-T is used to produce ASR hypothesis.

    \vspace{-1mm}
\section{Architecture Details}

    \vspace{-1mm}
\textbf{Standard Maestro}
We follow the architecture in~\cite{chen2022maestro} as the standard Maestro model for aligned speech-text representation learning.

\myparagraph{Speech encoder and shared encoder}: The speech and shared encoders are a stack of ``Conformer blocks''.  We use the Conformer XL architecture described in~\cite{chen2021injecting} with $24$ layers of full-context Conformer blocks (600M parameters) where the speech encoder comprises of the lower 6 layers and the upper 18 layers form the shared component, encoding both speech and text.

\myparagraph{Text encoder}: The text embedding extractor includes $512$ dimensional input embedding lookup layer,  3 convolutional layers of $512$ filters with kernel size $(5, 1)$, followed by a 6-layer Transformer with positional embedding. Durations for the injected text are modeled by repeating the original text embedding to the target length of  specified duration. 
%The upsampling is done by copying the original text embedding to the target length of  specified duration with positional embeddings to capture frame positions within text units as described in ~\cite{elias2021parallel}. %followed by positional embedding regarding to the frame positions within each text unit and its duration described in~\cite{elias2021parallel}.
%\andrew{absolute or relative position?}.  
%ZH: relative, edited 
The Refiner includes 2 layers of 8-headed self-attention blocks with $17\times 1$ lightweight convolutions~\cite{wu2019pay}. The duration model includes four blocks of $3\times 1$ lightweight convolutions transforming the original text embedding to predict the duration.  
%\andrew{what is the input and output of the duration model?} %ZH: added

%\andrew{weird to forward reference this.}
%ZH: done
\myparagraph{RNN-T decoder}: We use a 2-layer, 1280-dim LSTM with a joint network of 640 dims
as the RNN-T decoder. %While we use an RNN-T decoder in this paper, the proposed framework allows for the use of any decoder (CTC, LAS, etc.).
%\andrew{both?}
%zhehuai:done
By default we use grapheme as the target with the vocabulary size of 6100~\cite{conneau2022fleurs}. When using phoneme or byte
%or transliterated grapheme
as the target, we include an additional RNN-T decoder to predict to these and the original grapheme target independently.
%and VoxPopuli corpora by default (denoted as \textbf{\method{}}). When including mC4 dataset with 101 languages, we instead .
%\andrew{why is this denoted specially?}
%\zhehuai{try to differentiate the systems with phoneme+grapheme or spm-only}

\myparagraph{Additions in Maestro-U}
% 
%/cns/qu-d/home/zhehuai/brain/rs=6.3/M4MlsBabelVoxpMaestroFleursAaLidACleanTbAdapterBe.k/train/*params.txt
%LID
%decoder L emb
The language embedding in the input of the RNN-T decoder has 1024 dimensions and it is added to the original shared encoder output embedding. 
%text encoder L emb
The language embedding in the text encoder has 16 dimensions and it is concatenated with the text embedding extractor output.
%shared encoder: adapter
The language id based residual adapter~\cite{kannan2019large}  projects to a 32-dimensional bottleneck,  passes
through a ReLU non-linearity, then projects back up to the original
size. They are added after each of the conformer block in the shared encoder.

We explore different text encoder inputs to handle the unseen scripts while the grapheme RNN-T decoder is always used to produce ASR hypotheses.
%byte configuration
We use UTF-8 bytes~\cite{li2019bytes} as the text encoder input in the byte based experiments.
%G2P configuration
X-SAMPA is used in the phoneme experiment. 
%transliteration configuration
%The WFST based transliteration model described in~\cite{emond2018transliteration} is used in transliteration to latin experiments.

\myparagraph{Training hyper-parameters}
We include untranscribed speech, unspoken text, transcribed speech in each batch with fixed batch sizes, (1024, 8192, 512) respectively.
%To stabilize training, (1) We use exponential-moving-averaged (EMA) with decay rate $0.9999$ to stabilize prediction steps in training, including $\text{Align}_{\tt Rnnt}$ in transcribed speech, $\theta_{\tt Duration}$, $\text{Resample}$, and $\theta_\text{Refiner}$ in unspoken text when calculating $\mathscr{L}_\text{MoMa}$ and $\mathscr{L}_\text{A-MLM}$. 
%For fine-tuning, EMA variable was used to initialize downstream tasks. 
To follow the pretrain setup in~\cite{chen2022maestro} where exponential-moving-averaged (EMA) with decay rate $0.9999$ is used. 
A curriculum learning schedule starts from untranscribed speech-only training,  includes transcribed speech after 500k steps and unspoken text after another 15k steps. 
The joint training of three types of data lasts for another 300K steps  with a learning rate schedule and optimizer given in~\cite{zhang2020pushing}.    
%i) We keep a separate copy of exponential-moving-averaged model weights aggregated  and use it for $\text{Align}_{\tt Rnnt}$ in transcribed speech, $\theta_{\tt Duration}$, $\text{Resample}$, and $\theta_\text{Refiner}$ in unspoken text when calculating $\mathscr{L}_\text{MoMa}$ and $\mathscr{L}_\text{A-MLM}$. ii) We introduce a curriculum learning schedule to start from  untranscribed speech-only training,  include transcribed speech after 500k steps and unspoken text after another 15k steps. %AR is this schedule the same or different from the schedule for learning the aligner, etc?  can these be unified? so we start with untrasncribed speech-only, then include transcribed speech and train the aligner, (then train the duration model, and the matching model), finally include the extra text.
% zhehuai: 3 stages in total as mentioned above. no other more fine-grain schedule
%optimization

\myparagraph{Optional supervised finetuning}
In zero supervised speech ASR, we directly evaluate Maestro using its RNN-T decoder output as no fine-tuning is possible. For the other settings, we use the available transcribed data to conduct supervised multilingual fine-tuning.  All fine-tuning parameters follow ~\cite{chen2022maestro}. Since Maestro training incorporates both an encoder and decoder, in fine tuning experiments, both components are pretrained.

\begin{figure*}[tb!]
  \centering
  %https://docs.google.com/presentation/d/14VAZ1g9fk3e26dcrThMQFh2FxJmJ7K3AYSmGnfkrZ_Q/edit?resourcekey=0-TNlMztcZ0keeDZYOR4Bdzg#slide=id.g13e131f272c_0_0
  %https://docs.google.com/spreadsheets/d/1Yf1W2Fn1XMpqHUKtvzcR98Iwaz8JfCtaYpeOM23TYms/edit?resourcekey=0-ysiRsRpgCDnjZn5x1-GdUA#gid=340952447
\includegraphics[width=0.9\linewidth]{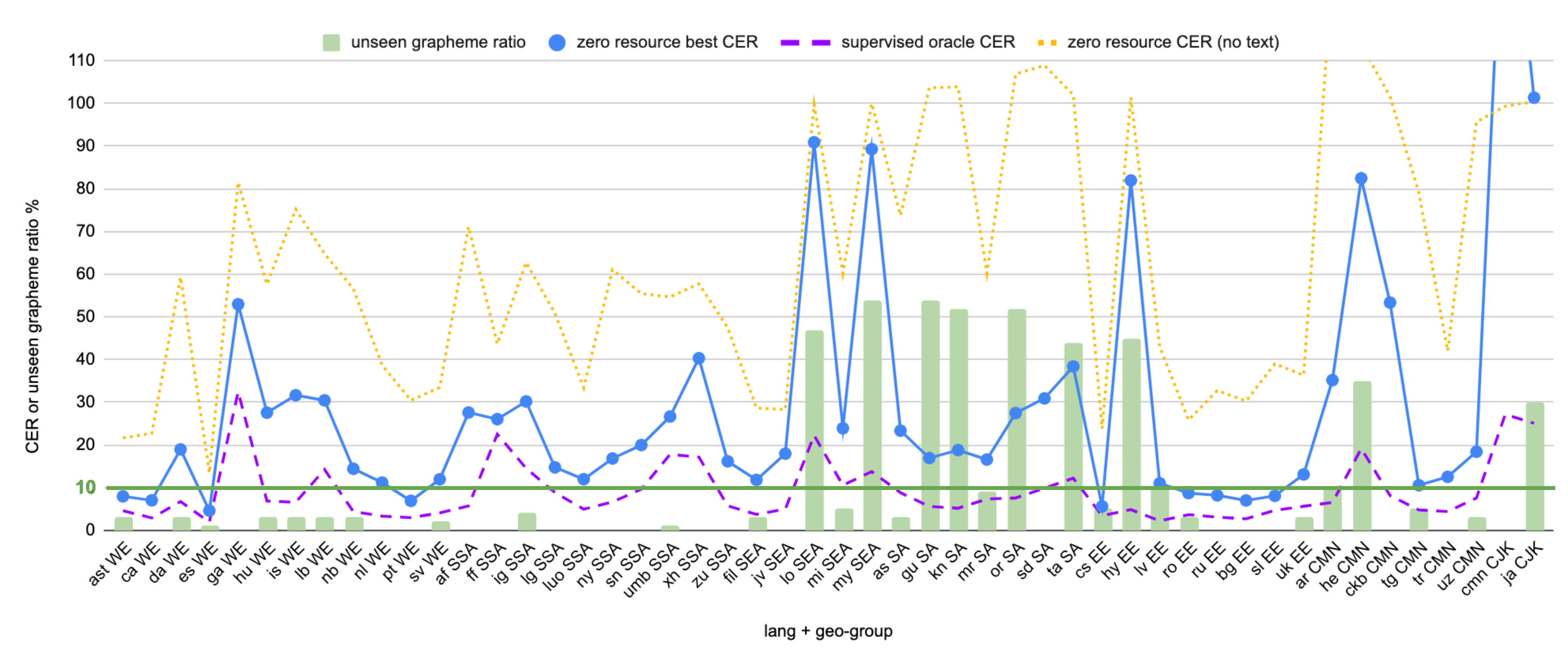}
    \vspace{-2mm}
    \caption{Per-language  breakdown of CER reduction with Maestro-U %in comparison with speech-only zero-supervised-speech baseline
    }
    \label{fig:breakdown_oracle}
    \vspace{-3mm}
\end{figure*}

    \vspace{-1mm}
\section{Experiments and Results}
    \vspace{-1mm}
%Will convert tables from this doc \url{http://shortn/_85fFAP1mzC}

\subsection{Overall Performance}
\label{sec:exp-overall}

The experiments in this paper fall under two scenarios: with supervised training data (manually labeled) and with zero supervised speech. Both scenarios include any available untranscribed speech and unspoken text in all the languages.
Table~\ref{tab:exp-all} summarizes the results under these two scenarios. In the {\it Supervised Setting} section of the table, the models are trained using all available supervised and unsupervised speech and unspoken text data from all the 102 languages, thus serving as our {\it oracle}. Column {\it A+B} presents the average performance of various models on all the 102 languages in the FLEURS corpus, while columns {\it A} and {\it B} denote performance on languages in group A and B respectively.
%models 
%trained 
%evaluated % modified by ZH
%with data from languages in group A and B respectively. 
We evaluate performance using Character Error Rate (CER) as proposed in~\cite{conneau2022fleurs}. We use a {\it W2V-BERT} model trained with untranscribed speech and 
fine-tuned with % modified by ZH
transcribed speech as the baseline for all experiments with no other text injection. %The proposed Maestro method with all the improvements described in Section~\ref{sec:mlang-design} reduces the CER of W2v-BERT baseline by 31.5\%.

Our first observation is when supervised data is available, the proposed method (Section~\ref{sec:mlang-design}) is able to reduce the average CER for all language groups, showing an overall relative improvements of  29.2\% for all 102 languages. A breakdown of the gains across language groups shows a relative win of 33.9\% and 24\% for languages in the groups {\it A}  and {\it B} respectively. 

The second half of Table~\ref{tab:exp-all} summarizes the performance of various models when zero supervised speech training material is available. The goal of this work is to bring the performance on the 50 languages in Group B as close as possible to Row 2 when supervised training data is available. We observe that the {\it W2V-BERT} model's performance worsens across all datasets when the supervised material from 
Group B languages is no longer available, with Group B languages degrading significantly by nearly 4  times. Models {\it W2V-BERT} and {\it Joint W2V-BERT} perform very similar to each other on the languages in Group B while on Group A languages, {\it Joint W2V-BERT} is significantly worse. The difference between these two models is that the former is fine-tuned on supervised data from Group A languages, while the latter eliminates fine-tuning by including a supervised RNN-T loss in the pretraining stage of {\it W2v-BERT}, a design previously explored in ~\cite{tts4pretrain2,icassp2022dongseong,bai2022just}. 
%\textbf{WHYYYYY}
% TODO: explain in catastrophic forgetting
%We build two none-text baseline as the first and second rows of the second section. The former uses the available group A supervised data to finetune the speech-only W2v-BERT pretrain model and the latter gets rid of the finetune by including joint supervised RNN-T loss into the W2v-BERT pretrain, similar to~\cite{bai2022just}.
%It shows that supervised finetune  using the available group A transcribed data slightly hurts  the performance on group B although it can improve the performance on group A.
%Similar to above, we did't see extra benefit on group B using supervised finetune  from the available group A transcribed data.
%On the other hand, 
%there is some quality difference between with and without fine-tune in the supervised setting above, even all the transcribed data has been used in the Maestro training, 
%M4MlsBabelVoxpMaestroFleursLidACleanTbMAdapterBe.k
%e.g. 10.4 v.s. 18.7  on the  \textbf{B} column  of the proposed Maestro.
%If always comparing the none fine-tune numbers, the final zero-supervised-speech system closes the gap by
%Future direction includes how to catch up the improvement from supervised fine-tune for the zero-supervised-speech setting while we exclude supervised fine-tune below.

The last two rows in the table clearly demonstrate the value of text injection.
The proposed Maestro-U method  significantly improves over both a speech-only baseline {\it W2V-BERT} and the text-injection method proposed in~\cite{chen2022maestro}, referred to as {\it Standard Maestro} in this table. The proposed method in the supervised setting yields the best possible performance ({\em oracle}) with a 9.7\% average CER on Group B languages. With no text injection and no supervised speech for Group B languages, the best performance that can be obtained is an average CER of 64.8\% 
({\em no text baseline}) % modified by ZH
. Injecting text using Maestro-U results in an average CER of 30.8\%, closing the gap to oracle performance significantly by 68.5\% relative. %30.8-9.7 Figure~\ref{fig:breakdown_oracle} illustrates the per-language wins of the proposed method compared to {\it W2v-BERT} under the zero supervised training data setting and the oracle CER.

\begin{table}[tbh]
\centering
 \caption{\label{tab:exp-indic} {Average CER Performance of various models on South Asian Languages from Group A and B. } }
\begin{tabular}{l|rr}
\toprule
method                                              & \multicolumn{1}{l}{SA Group A} & \multicolumn{1}{l}{\textbf{SA Group B}} \\
 \midrule
 %(Lid, adapter, updated text, loss scale) baseline
 %Indic2-nosp2.M4MlsBabelVoxpMaestroIndicALidACleanTbAdapter.l-00043000 0.589833 6 0.0972571 7 10
%Grapheme Maestro  & 9.7                                     & 56.3                                    \\
Grapheme Maestro  & 9.7                                     & 59.0                                    \\
%Indic2-nosp2.M4MlsBabelVoxpMaestroIndicALidACleanTbAdapterPhnb.l-00039000 0.141333 6 0.0893 7 10
\ \ + G2P                                                     & 8.9                                    & 14.1                                    \\
%\ \ + G2P                                                     & 8.9                                    & 13.6                                    \\
%Indic2-nosp2.M4MlsBabelVoxpMaestroIndicALidACleanTbAdapterBe.l-00037000 0.182967 6 0.0946429 7 10
\ \ + byte                                                    & 9.4                                     & 18.3                                    \\
%\ \ + byte                                                    & 9.4                                     & 17.2                                    \\
%Indic2-nosp2.M4MlsBabelVoxpMaestroIndicALidACleanTbAdapterTrans.l-00050000 0.692233 6 0.1224 7 10
%\ \ + transliteration                                         & 12.2                                    & 69.2                                    \\
%\ \ + transliteration                                         & 11.4                                    & 66.5                                    \\
 \midrule
 %Indic2-nosp2.M4MlsBabelVoxpMaestroFleurs3LidACleanTbMAdapter.k-00073000 0.116433 6 0.111857 7 13
Oracle maestro                                            & 11.2                                    & 11.6                                    \\
%Oracle maestro                                            & 10.3                                    & 11.8                                    \\
%Indic2-nosp2.M4MlsBabelVoxpMaestroIndicALidACleanTbAdapterNoText.l-00045000 1.00063 6 0.136986 7 10
No text baseline                                          & 13.7                                    & 100.0                                    \\
%No text baseline                                          & 10.4                                    & 99.4                                    \\
 \bottomrule
 
\end{tabular}
    \vspace{-4mm}
\end{table}

\subsection{Analysis of Maestro-U on unseen writing systems}

We analyze the behavior of the proposed model using a subset of the 13 South Asian languages (SAs) with very little to no graphemic overlap with the target languages (Note that the presence of some latin script is unavoidable due to code-switching in these target languages). %We selected SAs from group A and B in Section~\ref{sec:benchmark} %with the exception of Sindhi which we do not have proper G2P and Transliteration system to compare with. 
% 0.05 of as graphemes are unseen in group A.
% 0.56 of gu graphemes are unseen in group A.
% 0.53 of kn graphemes are unseen in group A.
% 0.09 of mr graphemes are unseen in group A.
% 0.55 of or graphemes are unseen in group A.
% 0.45 of ta graphemes are unseen in group A.
%The average unseen grapheme ratio (defined in Section~\ref{sec:benchmark}) of the 6 languages with zero-supervised-speech is 37.2\% and ranges from 5-56\%.  
The  unseen grapheme ratio (defined in Section~\ref{sec:benchmark}) of the 6 languages with zero-supervised-speech ranges from 5\% to 56\% and the average is  37.2\%.  

\begin{figure}[hbt!]
  \centering
    \vspace{-4mm}
\includegraphics[width=\linewidth]{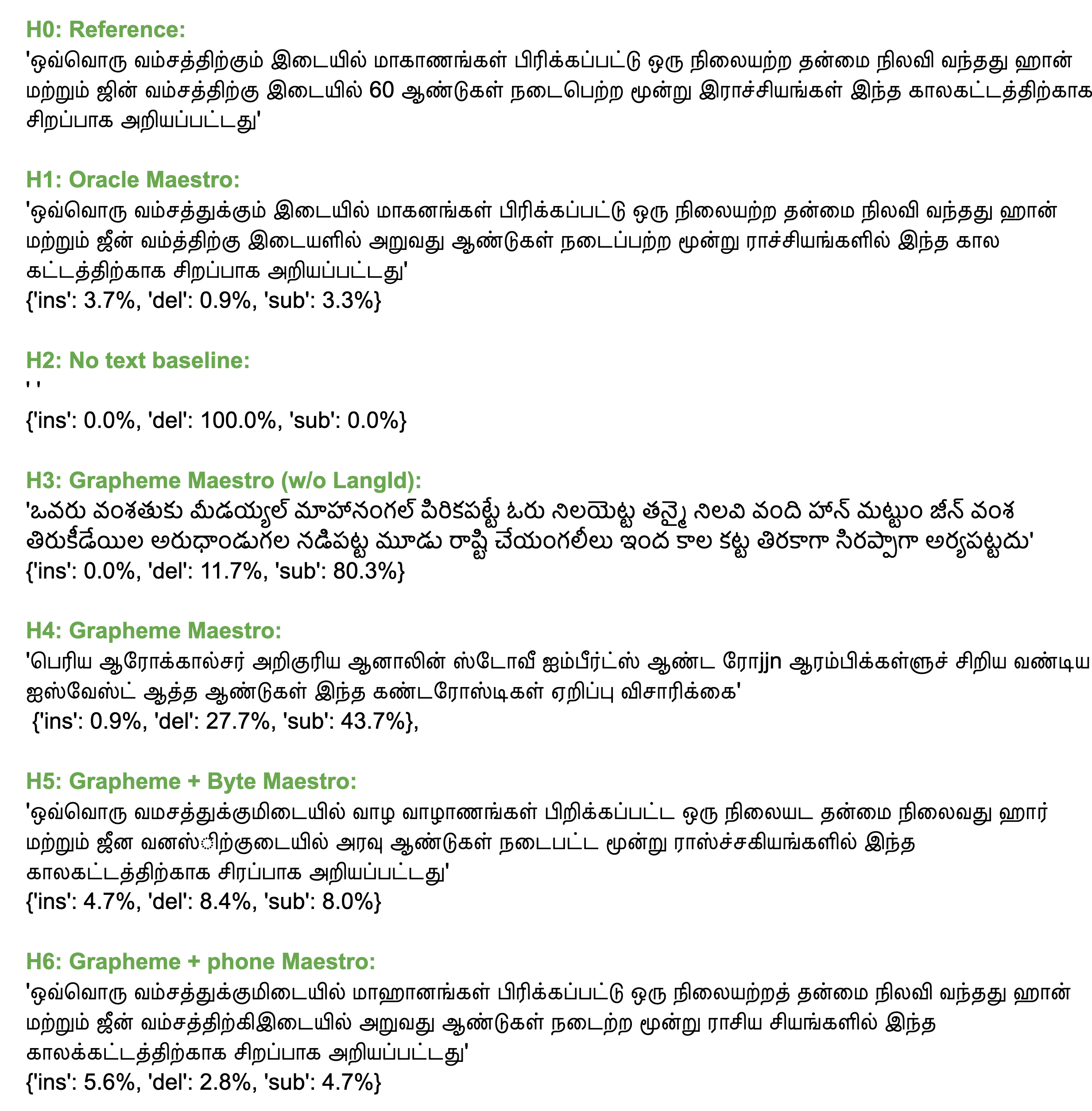}
    \vspace{-4mm}
  %https://docs.google.com/document/d/1v3yI-RDSf7qVceyXlVlsxNY5UF_9crji6-fHWwJ13CE/edit?resourcekey=0-RL1YqYY8HfoOPm8QaUv00Q
    \caption{Examples on Tamil from the 13 South Asian languages experiment
    }
    \label{fig:example}
    \vspace{-2mm}
\end{figure}

%https://docs.google.com/document/d/1v3yI-RDSf7qVceyXlVlsxNY5UF_9crji6-fHWwJ13CE/edit?resourcekey=0-RL1YqYY8HfoOPm8QaUv00Q
%To understand better, we fristly draw some example hypotheses  in 
Figure~\ref{fig:example} illustrates the behavior of several text injection methods using an utterance from {\it Tamil}. When the model is trained only with untranscribed speech %and the associated language embedding
from the set of Group B languages but without any form of text injection, there is no hypotheses produced during inference in the writing system of the target language (H2). In this extreme case, the model produces no output as it is impossible to learn a correlation between the writing system and the speech signal.  Meanwhile, when the model is free to produce a hypothesis unconstrained by a language embedding, it babbles in the closest {\it sounding} language seen in the training data (H3). 

With a duration model trained on graphemes  from Group A languages (H4), the model is able to output graphemes in the target language. Graphemes in the target language are hypothesized only when the target language's embedding is available to the model during training and inference. 
% begin of rewrite
%(ZH: try to rewrite above two sentences, WDYT?) With a duration model trained on graphemes  from Group A languages and the language embeddings (the fourth hypothesis), the model is usually able to pick up the correct script on a zero-supervised-speech language during inference. 
% end of rewrite
However, the model can still output graphemes from random languages and even form invalid words in the target language using graphemes there. 
This suggests that the model is unable to learn from the multilingual information shared through similar sounding graphemes across languages.

Clustering is a mechanism that can be used to explicitly encourage sharing of information across languages. This paper explores three methods: clustering graphemes with similar pronunications with the aid of a lexicon (H6), a simpler and deterministic grouping using the UTF-8 byte representations of the graphemes (H5).
%and a natural grouping by transliterating all the writing systems to one canonical (for example, Latin) writing system. 
It can be seen that these two clustering methods output graphemes with more acceptable error rates in the target language.
%In this Tamil example, the model without text injection produces empty hypothesis  as the textual form of this zero-supervised-speech language has never been learnt by the model even there is W2v-BERT based self-supervised encoder representation learning. With the text injection but without injecting language information, the model usually produces a hypothesis with certain wrong language seen in the supervised training data, 
% google translate says it's Telugu but pls help confirm
%Telugu in this example. 
%With language information, the grapheme based Maestro on the unseen language is still bad as there is only Lang Id that is forcing the model to spew some words in the Tamil script but many of them are not phonetically similar. This is due to the fact that the model fails to cluster these Tamil words learnt from unspoken text with the speech representations learnt from supervised data in other languages once there is no graphemeic overlap.
%To promote sharing, G2P can cluster  similar phonetic sounds across languages with human knowledge. Similarly but without extra knowledge, we propose to use UTF-8 byte encoding as the simplest deterministic way to project unseen graphemes to a sequence of 256-dim shared embeddings and make the model learn to cluster similar sounds in the Maestro training, which shows  similar effect on the example.

Table~\ref{tab:exp-indic} compares the error rates for the aforementioned methods. 
%As discussed in Section~\ref{sec:unseen}, G2P and Transliteration  both require extra human knowledge while byte decomposition based solution not. 
Both phonemic (G2P) and byte based modeling of durations can  significantly improve the performance on Group B languages with zero supervised speech. While byte-based encoding is slightly worse (18.3\% CER) than phonemic-based encoding (14.1\% CER), it is an extremely viable alternative for use in expanding to new languages previously unseen by the model with no available lexicon.
%%added for camera-ready - begin
We have two hypotheses as to why replacing graphemes with bytes would have such a dramatic effect on zero-supervised ASR: (i) 
%The design of Unicode enables this generalization. 
Unicode mappings from characters to bytes are often designed to encode similar characters in the same place across scripts. In the specific case of South Asian languages where characters correlate strongly with pronunciations, using a shared mapping in the text encoder might make it easier for the model to learn a more language-agnostic mapping.% across acoustics, bytes and graphemes. 
(ii) Using a shared byte-level vocabulary enables better generalization within the duration upsampler. 
%to upsample text representations to mimic the outputs of the speech encoder. 
This component is trained on alignments from the RNN-T decoder on supervised ASR data from non-zero-supervised languages. 
%It is unclear how well this upsampler could generalize to a completely different script. 
We hypothesize that sharing the text vocabulary in the form of bytes across languages 
%enables better generalization within the text encoder,
results in better aligned representations here for zero-supervised languages. 
%%added for camera-ready - end

%% bhuv add the thresholded cer based num of unseen langs improving here or later under analysis and conclusions
%Byte is still worse than G2P but requires no extra human knowledge. This indicates that when expanding to a new language with zero-supervised-speech data, if there is phoneme annotation, we should consider using it to get the best result. But if not, byte decomposition can still be a very good alternative. 
% fix the transliteration bug

Overall, 
we believe this is the first demonstration that competitive ASR performance can be achieved for an unseen language using no language resources other than text and untranscribed speech.
Byte-based clustered representations achieves a good balance between efficiency and scalability across languages with a language-wise performance presented in Figure~\ref{fig:breakdown_oracle}. %Note this was breakdown_byte
 When speech and text representations are  learnt in this manner, the model is able to recover well from the heavy losses resulting from a high unseen grapheme ratio. We hypothesize that representations learnt in this manner could serve as a foundation model for massively multilingual ASR.

% TODO: Andrew: I think we can drop all of the previous para but keep the last sentence. and also drop the figure below.

%Without the introduced byte solution, text injection usually results in much less improvement on those languages with relatively high unseen grapheme ratio. The introduced byte solution effectively recovers the loss from high unseen grapheme ratio and resultnosp2.M4MlsBabelVoxpMaestroFleursAaLidACleanTbAdapterBe.k-00107000 0.30866 50 0.211365 102 0.117812 52 13s in significant improvement from text injection on most of the languages.

%\begin{figure*}[hbt!]
%  \centering
%\includegraphics[width=0.8\linewidth]{fig/breakdown_byte.png}
%  %https://docs.google.com/spreadsheets/d/1Yf1W2Fn1XMpqHUKtvzcR98Iwaz8JfCtaYpeOM23TYms/edit?resourcekey=0-ysiRsRpgCDnjZn5x1-GdUA#gid=340952447
%  %NoTextMlsBabelVoxpMaestroFleursA.i
%  %M4MlsBabelVoxpMaestroFleursALidACleanTbMAdapter.k
%  %M4MlsBabelVoxpMaestroFleursA3LidACleanTbAdapterBe.k
%    \caption{Per-language CER breakdown of the improvement from byte %decomposition
%    }
%    \label{fig:breakdown_byte}
%\end{figure*}

   \vspace{-1mm}
\subsection{Ablation study}
\label{sec:exp-ablation}
   \vspace{-0.5mm}

In this section, we measure the impact of the Maestro-U design choices described in Section~\ref{sec:maestro-u}.
Table~\ref{tab:exp-ablation-proposed} presents an ablation study of the various components of Maestro-U under the zero-supervised-speech setting. While each of the proposed modifications provide gains, scaling up the text loss (5.4\%), use of adapters (6.9\% ), byte-based encoding(10.3\%), and use of in-domain text injection (13.3\%) offer the maximum relative wins.
%All the improvements contribute to the final win.
\begin{table}[tbh]
\centering
\caption{\label{tab:exp-ablation-proposed} {Ablation study of the proposed method. } }
\begin{tabular}{l|l}
\toprule
method                              & CER on group B \\
\midrule
no text baseline                    & 64.1                   \\
\midrule
%nosp2.MC4MlsBabelVoxpMaestroFleursA.i-00037000 0.542154 50 0.395733 102 0.254944 52 11
Standard Maestro~\cite{chen2022maestro} & 54.2                   \\
%nosp2.M4MlsBabelVoxpMaestroFleursALidACleanT.k-00091000 0.48531 50 0.337777 102 0.195919 52 11
\ \ + Language-id                               & 48.5                   \\
%nosp2.M4MlsBabelVoxpMaestroFleursALidACleanTbWb.k-00167000 0.458824 50 0.295223 102 0.137913 52 11
\ \ \ \ + Upscale text loss                 & 45.9                   \\
%nosp2.M4MlsBabelVoxpMaestroFleursALidACleanTbMAdapter.k-00088000 0.427154 50 0.282075 102 0.142577 52 12
\ \ \ \ \ \ + Adapter                           & 42.7                   \\
%nosp2.M4MlsBabelVoxpMaestroFleursALidACleanTbAdapterBe.k-00219000 0.383558 50 0.248237 102 0.118121 52 12
\ \ \ \ \ \ \ \ + Byte                              & 38.3                   \\
%nosp2.M4MlsBabelVoxpMaestroFleursA6LidACleanTbAdapterBe.k-00119000 0.331998 50 0.229969 102 0.131863 52 11
\ \ \ \ \ \ \ \ \ \ + In-domain text                & 33.2                   \\
%decay w2b loss
%nosp2.M4MlsBabelVoxpMaestroFleursAaLidACleanTbAdapterBe.k-00107000 0.30812 50 0.211365 102 0.117812 52 13
\ \ \ \ \ \ \ \ \ \ \ \ +  Loss tapering            & 30.8                  \\
\bottomrule
\end{tabular}
\end{table}
%To demonstrate the necessity of aligned speech-text representation learning, 
Next, we conduct an ablation study on the Maestro text encoder in Table~\ref{tab:exp-ablation-maestro}.
%Experiment result and analysis
Specifically, we investigate the value of 1) text resampling to match the speech frame rate 2) training a duration model, 3) consistency loss between text and speech, 4) whether to use the duration model at inference. %Results are reported .
% The ablation is conducted on encoder architectures for deriving text representations, training losses used to train the text encoder and a learnt duration model. 

When the duration model is not used during text based training, we either sample durations uniformly from one to four frames or use one frame. In~\cite{chen2022maestro}, the RNN-T decoder loss is not back-propagated to the text encoder. 
%In this work, the text encoder is always trained by back-propagating the RNN-T decoder loss from the unspoken text similar to~\cite{bapna2022mslam}. 
% ZH: the above sentence is not correct, replace it with the following.
%BR I do not follow this sentence. If you are training a duration model you need a rnnt loss, is it not? Consistency is optional.
In this work, when consistency loss is not used, the text encoder is trained by back-propagating the RNN-T decoder loss from the unspoken text similar to~\cite{bapna2022mslam}. % and \cite{tts4pretrain2}. 
% The first four columns in Table~\ref{tab:exp-ablation-maestro} evaluate the role of resampling and refining within the text encoder, need for learning duration from alignments, consistency loss and the durations used during text injection.
%
Note that even if there is a duration model learned during training, %to learn the shared speech-text representation, 
it is not necessary to  use this model during text-only training.  The second  row shows the impact of this.  We find that learning aligned speech-text representations is crucial for text injection to yield gains in ASR, while the impact of the duration model during text-only training is more mixed.

\begin{table}[tbh]
\centering
\caption{\label{tab:exp-ablation-maestro} {Ablation study of the text encoder using grapheme based Maestro in the 5th row of Table~\ref{tab:exp-ablation-proposed}. } }
\begin{tabular}{llll|rr}
\toprule
%Resamp-  & Trained   & Consi-  & Duration  & \multicolumn{1}{l}{A}  & \multicolumn{1}{l}{B} \\
% ling & Duration &  stency  &  Prediction & CER  &    \\
Resamp-  & Trained   & Consi-  & Duration  & \multicolumn{2}{c}{CER}     \\
 ling & Duration &  stency  &  Prediction &  \multicolumn{1}{c}{A} & \multicolumn{1}{c}{B}  \\
\midrule
% this row is the proposed grapheme based Maestro
%nosp2.M4MlsBabelVoxpMaestroFleursALidACleanTbMAdapter.k-00088000 0.427154 50 0.282075 102 0.142577 52 12
Y & Y & Y & Learnt          &14.2  & 42.7                               \\
%  r.e. 4) why duration prediction is worse than uniform on group A: here is the breakdown of the outlier better number in row 2
%so yes, duration prediction seems to work pretty bad on two langs, ko and sr which results in higher WER on group A https://screenshot.googleplex.com/AssyC4nL3MPVb2E
%nosp2.M4MlsBabelVoxpMaestroFleursALidACleanTbAdapterU.k-00084000 0.449946 50 0.289125 102 0.13449 52
Y & Y & Y & Uniform          &13.5   & 45.0                               \\
%nosp2.M4MlsBabelVoxpMaestroFleursALidACleanTbAdapterU3.k-00037000 0.587482 50 0.398064 102 0.215931 52 13
Y & Y & N & Learnt         &21.6    & 58.7                               \\
%nosp2.M4MlsBabelVoxpMaestroFleursALidACleanTbAdapterU2b.k-00052000 0.58045 50 0.368423 102 0.16455 52 15
Y & N & N & Uniform        &16.5    & 58.0                               \\
%nosp2.M4MlsBabelVoxpMaestroFleursALidACleanTbAdapterU4.k-00016000 0.7627 50 0.556204 102 0.35765 52 13
N & N & N & None                &35.8     & 76.3                                 \\
\bottomrule
\end{tabular}
\end{table}

\section{Conclusion}
\label{sec:conclu}
   \vspace{-0.5mm}

We propose the use of joint speech and text representation learning to develop ASR systems for languages with zero supervised training data.  Using a massively multilingual setting comprising of $102$ languages, of which $50$ languages have zero supervised training data, we show that Maestro-U is capable of multilingual knowledge transfer. On these $50$ languages, Maestro-U successfully brings the CER of \textbf{9 languages below 10\%} and \textbf{19 languages below 15\%} closing the gap to oracle performance by 68.5\% relative. We show  significant improvements  on languages with and without overlapping writing systems, thereby making Maestro-U a viable solution for expanding to new languages and serving as a foundation model for massively multilingual ASR.
\bibliographystyle{IEEEbib}
\bibliography{strings,refs}

\end{document}